\documentclass{article}

\usepackage{microtype}
\usepackage{footnote}
\usepackage{graphicx}
\usepackage{subfigure}
\usepackage{booktabs} 
\usepackage{float}
\usepackage{hyperref}
\usepackage[numbers]{natbib}

\usepackage[accepted]{icml2025}

\usepackage{amssymb}
\usepackage{mathtools}
\usepackage{amsthm}
\usepackage{xcolor} 
\usepackage{wrapfig}
\usepackage{wrapfig}
\usepackage{multirow}
\usepackage{makecell}
\usepackage{subfigure}
\usepackage{placeins}
\usepackage{longtable}
\usepackage{setspace} 
\usepackage{stix} 
\usepackage[font=small,labelfont=it]{caption} 
\makeatletter
\renewcommand{\@makecaption}[2]{%
  \vskip 10pt
  \baselineskip 11pt
  \sbox\@tempboxa{\small\textit{#1.} #2}%
  \ifdim \wd\@tempboxa >\hsize
    \sbox\@tempboxa{\small\textit{#1.} }%
    \parbox[t]{\hsize}{\usebox\@tempboxa {\footnotesize #2}}%
  \else
    \centerline{\usebox\@tempboxa}%
  \fi}
\makeatother

\usepackage[capitalize,noabbrev]{cleveref}

\theoremstyle{plain}

\theoremstyle{definition}

\theoremstyle{remark}

\usepackage[textsize=tiny]{todonotes}

\newcommand{\data}{\textbf{Pensez-2k}}
\newcommand{\modelpensez}{\textbf{Pensez 7B}}
\newcommand{\modelqwen}{\textbf{Qwen2.5 7B Instruct}}
\newcommand{\modeldistil}{\textbf{DeepSeek R1 7B}}
\newcommand{\modelopenthinker}{\textbf{OpenThinker 7B}}
\newcommand{\modelopenhf}{\textbf{OpenR1 7B}}

\icmltitlerunning{\textbf{Pensez}:  Less Data, Better Reasoning – Rethinking French LLM}

\makeatletter
\AtBeginDocument{
\renewcommand{\sectionautorefname}{\S\@gobble}

\renewcommand{\subsectionautorefname}{\S\@gobble} 
\renewcommand{\subsubsectionautorefname}{\S\@gobble}
\renewcommand{\appendixautorefname}{\S\@gobble}

}
\makeatother

\definecolor{olmoeDarkYellow}{HTML}{fdac15}
\definecolor{defaultblue}{HTML}{0077B6}
\definecolor{defaultlightblue}{HTML}{00B4D8}
\definecolor{blue}{HTML}{03045E}
\definecolor{blueb}{HTML}{0077B6}
\definecolor{bluec}{HTML}{00B4D8}
\definecolor{blued}{HTML}{90E0EF}
\definecolor{bluee}{HTML}{CAF0F8}

\makeatletter
\let\@oldmaketitle\@maketitle
\renewcommand{\@maketitle}{\@oldmaketitle
  \includegraphics[width=\linewidth,height=4\baselineskip]
    {scaling}\bigskip}
\makeatother

\begin{document}

\twocolumn[{
\icmltitle{\textbf{Pensez}: Less Data, Better Reasoning – Rethinking French LLM}

\icmlsetsymbol{equal}{*}

\begin{icmlauthorlist}
\icmlauthor{Huy Hoang Ha}{m,u}
\end{icmlauthorlist}

\icmlaffiliation{m}{Menlo Research}
\icmlaffiliation{u}{Université Grenoble Alpes}


\icmlkeywords{Machine Learning, Large language models, Data curation, Test-time compute}

\vskip 0.3in
}]

\printAffiliationsAndNotice{}

\begin{abstract}
Large language models (LLMs) have demonstrated remarkable capabilities in various natural language processing tasks.
However, achieving strong performance in specialized domains like mathematical reasoning and non-English languages often requires extensive training on massive datasets.
This paper investigates a contrasting approach: strategic fine-tuning on a small, high-quality, bilingual (English-French) dataset to enhance both the reasoning capabilities and French language proficiency of a large language model.
Rather than relying on scale, we explore the hypothesis that targeted data curation and optimized training can achieve competitive, or even superior, performance.
We demonstrate, through targeted supervised fine-tuning (SFT) on only 2,000 carefully selected samples, significant improvements in mathematical reasoning.
Specifically, \modelpensez{} exhibits an increase in accuracy of the base model up to 20 points on the AIME25 and a 12 points increase on a French MATH level 5 benchmark.
These results challenge the prevailing assumption that massive datasets are a prerequisite for strong reasoning performance in LLMs, highlighting the potential of strategic data curation and optimized fine-tuning for enhancing both specialized skills and multilingual capabilities.
Our findings have implications for the efficient development of high-performing, multilingual LLMs, especially in resource-constrained scenarios.

\end{abstract}

\section{Introduction}
\label{sec:intro}

\begin{figure}
\centering
\includegraphics[width=0.95\columnwidth]{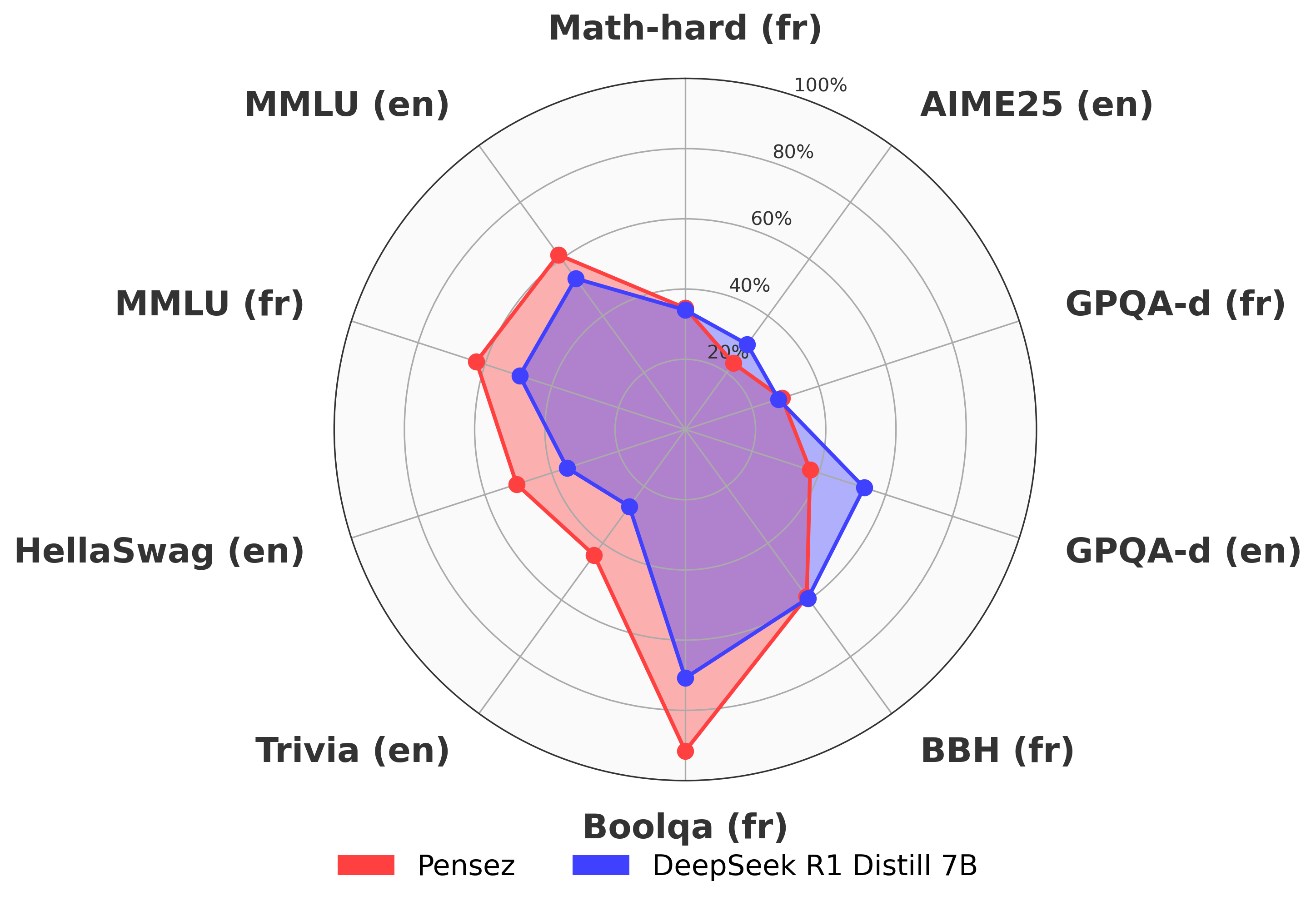}
\caption{\textbf{Performance on knowledge retrieval and reasoning skills in French and English}. Despite fewer training samples,  \modelpensez{} performs comparably to \modeldistil{} on reasoning tasks but outperforms it in knowledge retrieval.}
\label{fig:compare}
\end{figure}

Large language models (LLMs) have fundamentally reshaped the landscape of natural language processing (NLP), achieving unprecedented performance across a spectrum of tasks, from text generation and question answering to machine translation and code synthesis ~\cite{brown2020languagemodelsfewshotlearners, touvron2023llama, openai2023gpt4}. This progress, however, has been largely driven by scaling up model size and training data, leading to substantial computational costs and resource demands ~\cite{kaplan2020scalinglawsneurallanguage,hoffmann2022trainingcomputeoptimallargelanguage}. Furthermore, while LLMs demonstrate impressive fluency and broad knowledge, achieving robust performance in specialized domains requiring complex reasoning, such as mathematics ~\cite{lewkowycz2022solvingquantitativereasoningproblems, shao2024deepseekmathpushinglimitsmathematical, azerbayev2023llemma} and scientific problem-solving, remains a significant challenge.

Another persistent challenge is the disparity in performance across languages. Despite efforts to create multilingual LLMs ~\cite{conneau2018senteval,scao2022language}, most models are trained primarily on English-centric corpora, leading to a significant performance gap for many other languages ~\cite{martins2024eurollmmultilinguallanguagemodels, kudugunta2023madlad400multilingualdocumentlevellarge, lai2023okapiinstructiontunedlargelanguage}. Projects like EuroLLM ~\cite{martins2024eurollmmultilinguallanguagemodels}, CroissantLLM ~\cite{faysse2024croissantllmtrulybilingualfrenchenglish}, Aya ~\cite{2024ayamodelinstructionfinetuned}, and mT5 ~\cite{xue2021mt5massivelymultilingualpretrained} aim to address this imbalance by developing LLMs with enhanced capabilities in a wider range of languages. While these models improve the understanding of multilingual, they lack of reasoning capability to solve chalenging tasks.

The conventional paradigm in LLM training assumes that massive datasets are indispensable for strong performance, especially in complex reasoning tasks. However, recent work is beginning to challenge this assumption, demonstrating the potential of data efficiency. While approaches like DeepSeek-R1 ~\cite{deepseekai2025deepseekr1incentivizingreasoningcapability} have achieved impressive reasoning performance, they still rely on extensive reinforcement learning with millions of samples. In contrast, LIMO ~\cite{ye2025limoreasoning} and S1 ~\cite{muennighoff2025s1simpletesttimescaling} highlight the surprising effectiveness of small, meticulously curated datasets for fine-tuning LLMs. These findings suggest that strategic data selection, coupled with techniques that encourage extended reasoning during both training and inference, such as chain-of-thought prompting ~\cite{wei2022finetuned}, and increased test-time compute ~\cite{yao2023reactsynergizingreasoningacting, snell2024scalingllmtesttimecompute}, can unlock significant performance gains with dramatically reduced data requirements.

This paper advances this emerging trend by investigating targeted supervised fine-tuning (SFT) on a small, high-quality, bilingual (English-French) dataset. We hypothesize that prioritizing data quality, diversity, balanced bilingual representation, and, crucially, the inclusion of detailed reasoning chains within the training data can achieve competitive, or even superior, performance compared to models trained on vastly larger, less curated datasets. Specifically, we posit that by providing explicit examples of step-by-step reasoning processes, akin to extending the computational "thinking time" of the model, we can significantly enhance both the reasoning capabilities and French language proficiency of an LLM. This approach not only challenges the necessity of massive datasets for strong reasoning but also has significant implications for the efficient development of high-performing, multilingual LLMs, especially in resource-constrained settings or for languages with limited data availability.

Our main contributions are as follows:

1. Demonstration of Data Efficiency: We demonstrate that strategic fine-tuning on a small, high-quality dataset (2,000 examples) can yield significant improvements in both general reasoning and French-specific tasks.

2. Bilingual Enhancement: We show that a balanced bilingual training approach (1:1 English-French ratio) enhances performance in both languages, addressing the common imbalance in multilingual LLMs.

3. Resource Availability: We release our curated dataset, training code, and fine-tuned model to facilitate reproducibility and further research in this area.

\begin{figure}
\centering
\includegraphics[width=0.95\columnwidth]{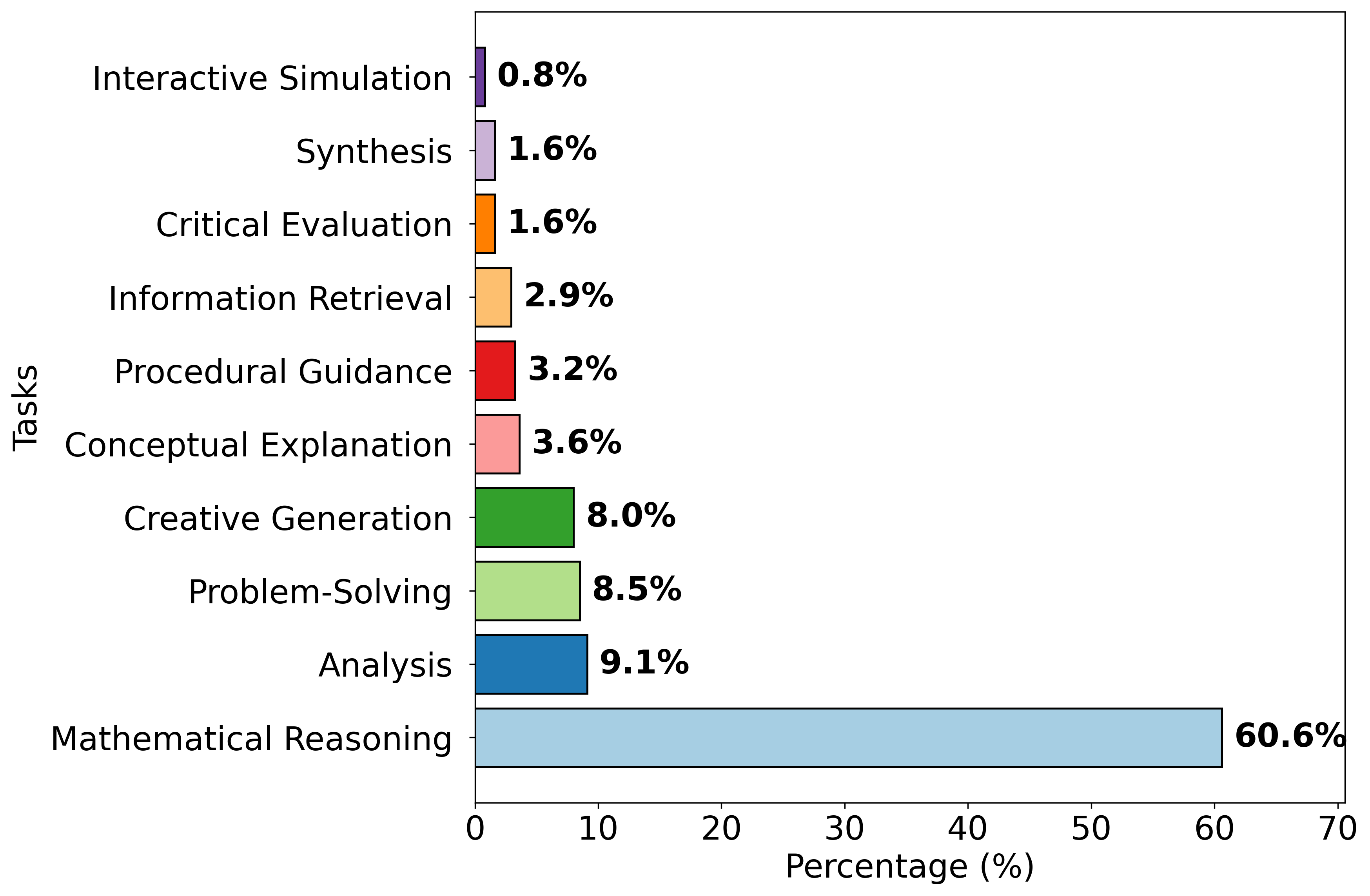}
\caption{The \data{} dataset is categorized into 10 prevalent task types reflecting typical user interactions.}
\label{fig:tasks}
\end{figure}

\section{Multilingual reasoning data \data{}}
\label{sec:data}

The development of a high-quality dataset is fundamental to advancing large language model (LLM) research, particularly under resource-constrained training conditions. This section details the methodology for curating and preprocessing datasets from reliable internet sources, emphasizing quality assurance and data integrity.

\subsection{Initial collection from existing datasets}
\label{sec:init-collection}

To construct a comprehensive training corpus, we collected datasets from publicly available, reliable sources, categorized into two main domains: reasoning and daily conversation.

\paragraph{Reasoning Datasets} We selected datasets that emphasize both depth and diversity in problem-solving. These include 817 carefully curated mathematical reasoning samples from \textbf{LIMO} ~\cite{ye2025limoreasoning}; the \textbf{Dolphin-r1} reasoning subset, which provides 300,000 reasoning traces across various domains ~\cite{2025dolphinr1}; \textbf{OpenR1-Math-220k}, containing 220,000 verified mathematical problems ~\cite{2025openR1math220k}; and \textbf{s1K-1.1}, featuring 1,000 curated challenging questions from 15 reliable mathematical question sources, with detailed reasoning chains to support test-time scaling ~\cite{muennighoff2025s1simpletesttimescaling}. Notably, datasets incorporating DeepSeek-R1 distillations serve as a robust foundation for structured problem-solving.

\paragraph{Daily Conversation Datasets} To encompass multi-turn, single-turn interactions, and long-prompt conversation requests, we selected the \textbf{Magpie} LLama-3-70B and Qwen 2.5-72B collections, which synthesize 300,000 high-quality pairs for alignment ~\cite{xu2024magpie}, as well as the training set of the \textbf{Tulu-3} model with 939,344 samples with diverse domain conversational data ~\cite{olmo20242olmo2furious}. These datasets balance scale and diversity, enabling LLMs to generate nuanced, context-aware dialogue. More importantly, they help mitigate biases introduced by mathematical reasoning-focused datasets.

\subsection{Data filtering}
\label{sec:selection-criteria}

Our objective is to construct an optimally compact and high-quality dataset that minimizes computational resource demands during model training while retaining linguistic robustness. To achieve this, we implement a three-stage filtering pipeline, systematically refining an initial corpus into a final set of 2,000 samples, evenly distributed across two languages-English and French (1,000 samples per language). This curated selection is guided by three principal criteria: \textit{Length}, \textit{Language Purity}, and \textit{Diversity}. These criteria are enforced following an initial quality assurance phase during data collection, ensuring that only high-fidelity samples proceed to subsequent refinement stages.

\paragraph{Length} Given that the target model is designed to handle a context length of 16,384 tokens, we filter out all samples exceeding this token count. Tokenization is performed using the tokenizer of the base model, \modelqwen{}, to ensure consistency with the training framework.

\paragraph{Language Purity} To verify linguistic integrity, we employ the FastText language identification model ~\cite{joulin2016fasttext}. For each language, we establish a confidence threshold of 0.95, ensuring that only samples with high linguistic homogeneity are retained.

\paragraph{Diversity} To quantify diversity, we categorize each question into distinct task types using the Llama 3.3 70B Instruct model. The dataset is deliberately biased toward reasoning-related tasks to strengthen the model's reasoning capabilities, while the remaining tasks are distributed evenly to maintain a balanced representation across other categories.

Following the application of these filters, our aim is to construct a final dataset comprising 2,000 samples, balanced evenly across English and French subsets (1,000 samples per language). However, due to an insufficient number of French samples—falling short of the target 1,000—we augment the French subset by randomly selecting additional samples from the English corpus and designating them as French for subsequent processing. This adjustment ensures the dataset meets the desired size and language distribution prior to the data transformation stage. Detailed sample counts for each subset, including the adjusted totals, are presented in \autoref{tab:dataset_breakdown}. All of the questions are deduplicated to ensure the diversity of the dataset.

\begin{figure*}[t]
\centering
\includegraphics[width=0.9\textwidth]{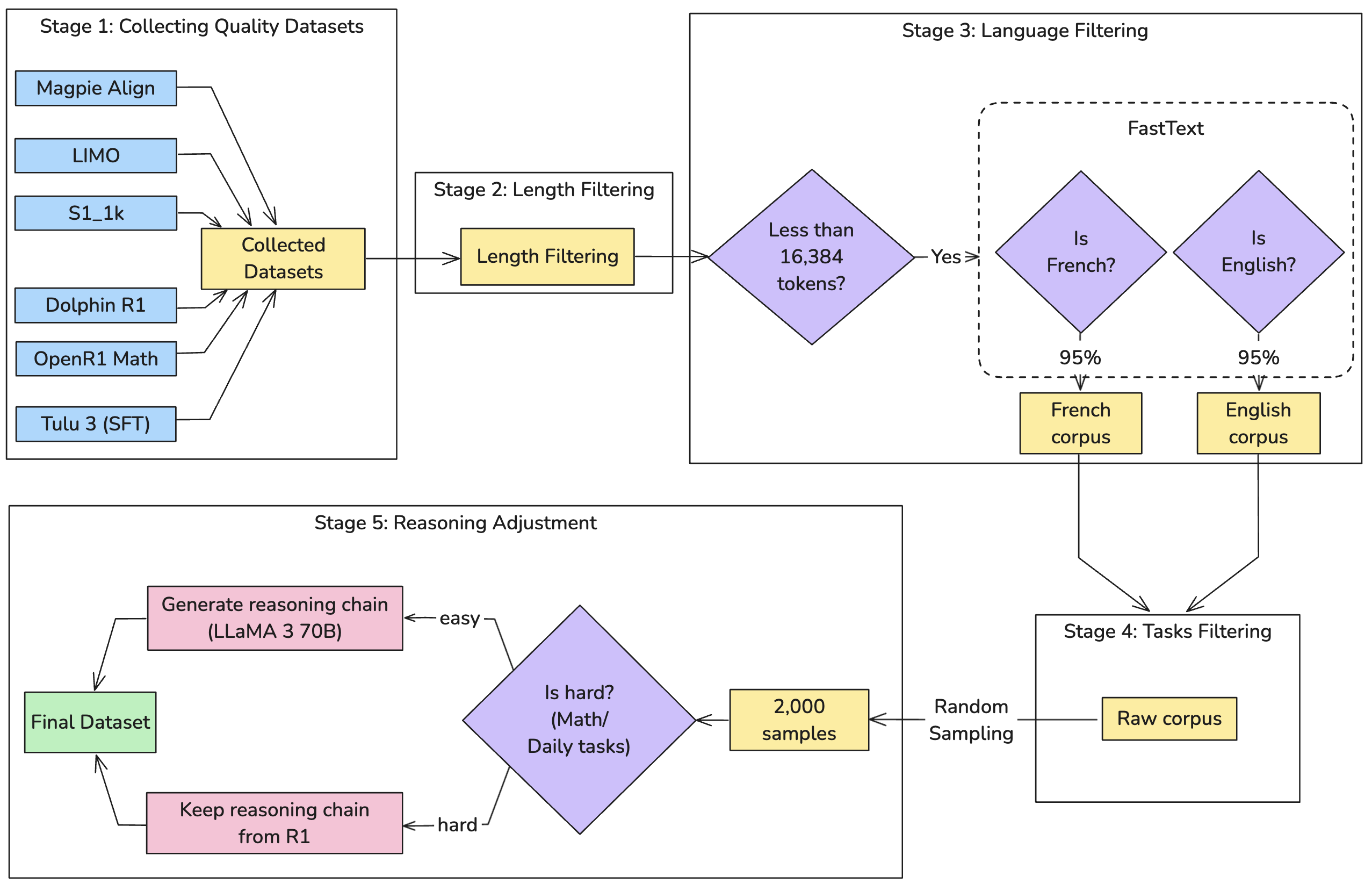}
\caption{\textbf{Data collection and cleaning pipeline for the \data{} dataset.} The pipeline consists of three stages: initial collection, data filtering, and data augmentation. The final dataset comprises 2,000 samples, evenly distributed across English and French, with a focus on diverse reasoning tasks.}
\label{fig:scaling2}
\end{figure*}

\subsection{Data augmentation}
\label{sec:data-augmentation}

To ensure a robust bilingual dataset capable of supporting advanced reasoning, we augment the curated samples through translation and enrich their reasoning content. This dual process addresses both linguistic balance and cognitive depth, preparing the dataset for effective model training. The following paragraphs detail the translation of samples across languages and the enhancement of reasoning chains. We used SGLang ~\cite{zheng2024sglangefficientexecutionstructured} framework to generate new augmented samples.

\paragraph{French Translation} To enable bilingual training across English and French, we translate the English samples into French using the Llama 3.3 70B Instruct model. This translation step ensures that the dataset maintains an equitable representation of both languages, with 1,000 samples per language. By aligning the linguistic content, we enhance the model to generalize reasoning capabilities across linguistic boundaries, enhancing its cross-lingual robustness. Prompt we used for translation is in \autoref{sec:translation}.

\paragraph{Reasoning Chain Augmentation} Samples derived from daily conversation datasets, typically comprising simple question-answer (Q\&A) pairs devoid of explicit reasoning, necessitate additional refinement to align with our training objectives. To this end, we utilize the Llama 3.3 70B Instruct model to generate concise reasoning chains for these samples. This augmentation ensures a uniform structural format across the dataset, seamlessly integrating tasks ranging from straightforward daily-life scenarios to complex reasoning challenges. Augmentation prompt is in \autoref{sec:augmentation}.

\section{Training reasoning model}
\label{sec:training}

We adopt \modelqwen{} as the base model for supervised fine-tuning on our \data{} dataset. To improve the model's ability to articulate its reasoning process, we introduce two special tokens, "<think>" and "</think>", which indicate the beginning and end of the reasoning steps. The training process employs full-parameter fine-tuning, optimized with DeepSpeed ZeRO-3 ~\cite{rasley2020deepspeed} and FlashAttention2 ~\cite{dao2023flashattention}, and is constrained to a maximum sequence length of 16,384 tokens. To enhance model robustness and generalization, we inject noise into the embedding layers during fine-tuning, following the NEFTune approach ~\cite{jain2023neftunenoisyembeddingsimprove}. Training is conducted on 8 NVIDIA H100 GPUs, with hyperparameters detailed in \autoref{sec:details-training}, completing in approximately 76 minutes.

\begin{figure*}[t]
\centering
\subfigure[Performance on English tasks]{\includegraphics[width=0.49\textwidth]{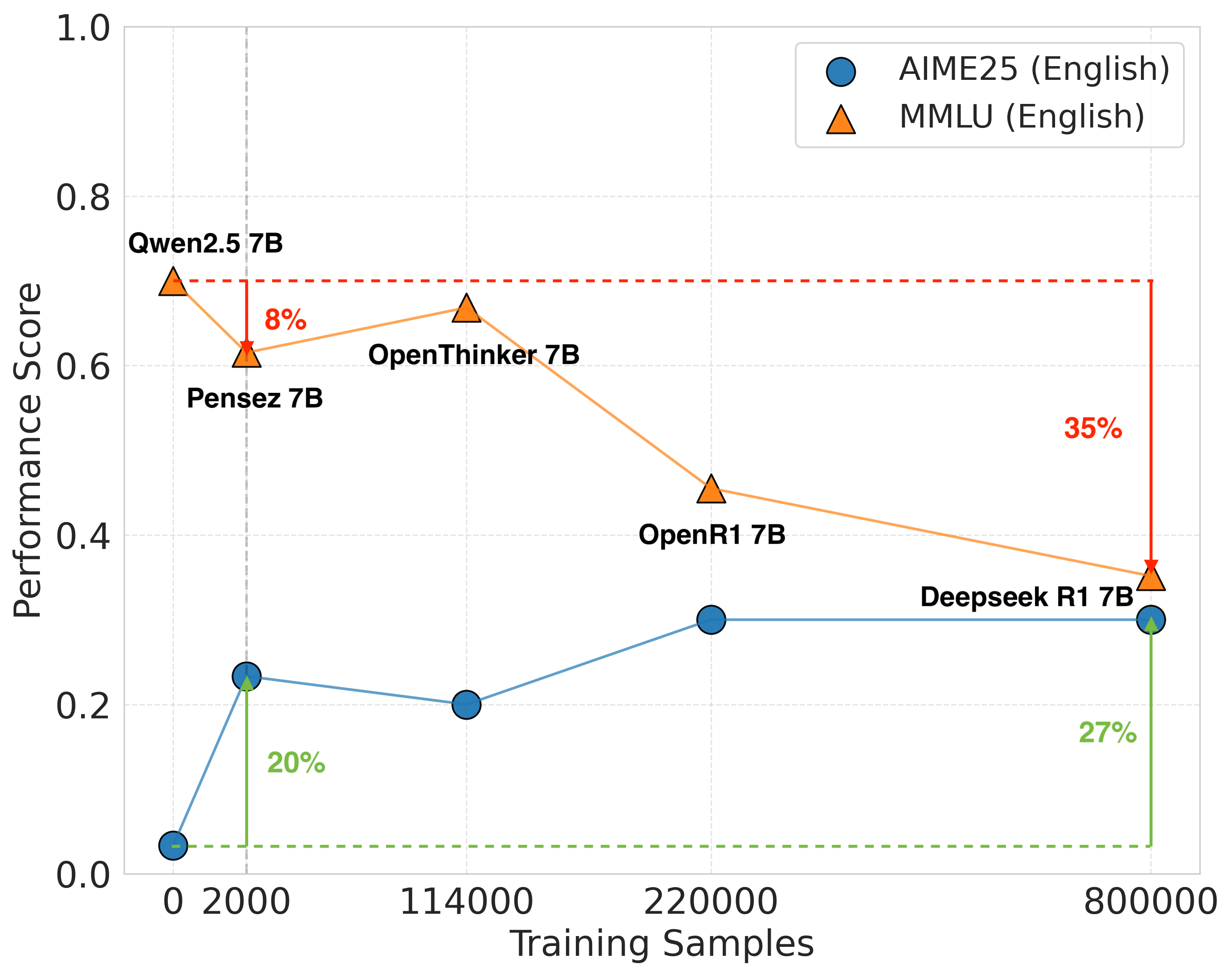}\label{fig:peft_en}}
\subfigure[Performance on French tasks]{\includegraphics[width=0.49\textwidth]{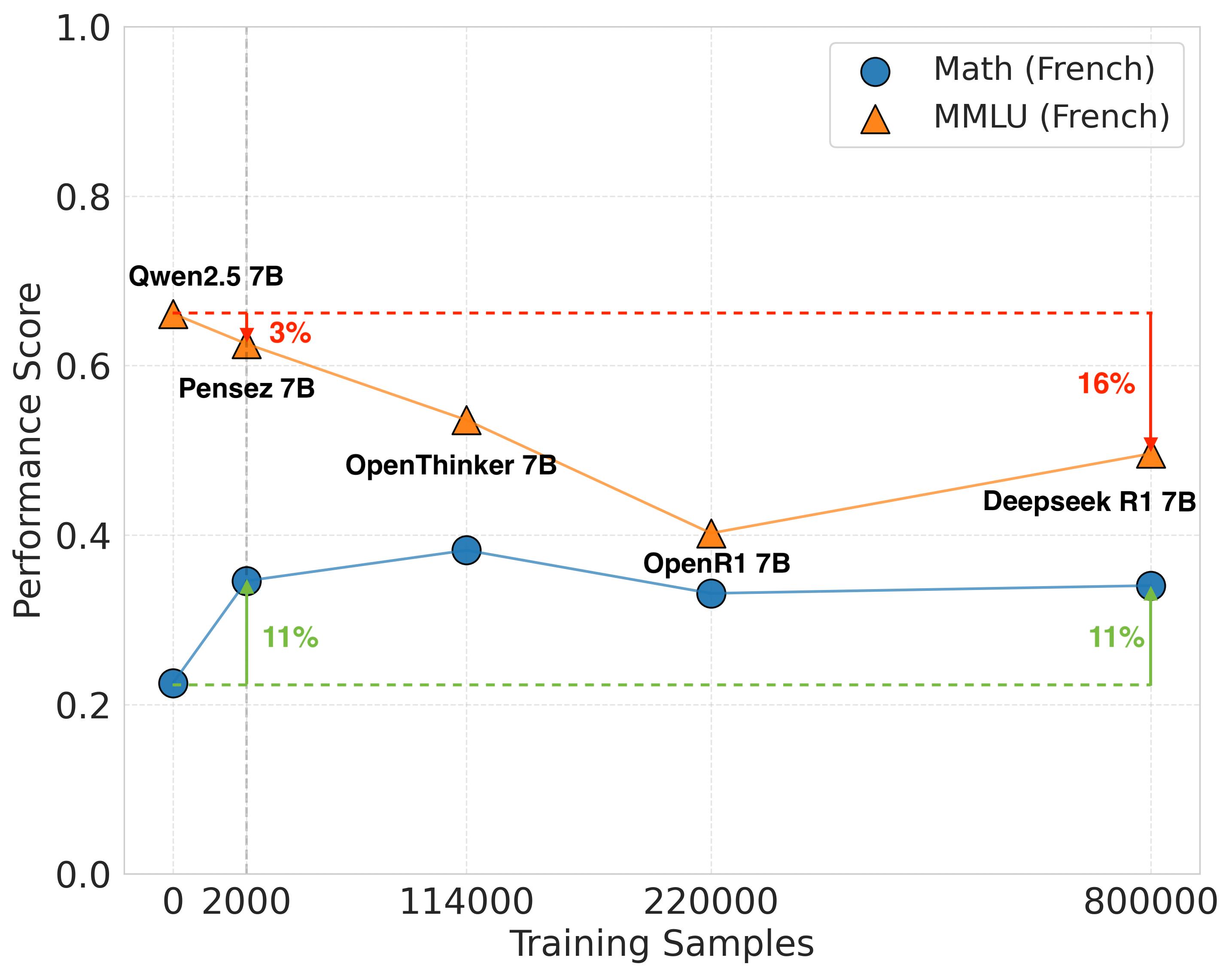}\label{fig:peft_fr}}
\caption{Comparative performance on (a) AIME25 and MMLU (English), and (b) MATH Hard lv5 and MMLU (French). All models are finetuned from the \modelqwen{} base model and undergo SFT with varying sample sizes. \modelpensez{}, trained on only 2,000 samples distributed across diverse tasks, achieves competitive performance on reasoning tasks while exhibiting minimal degradation in knowledge comprehension compared to other 7B reasoning models. This demonstrates its data efficiency relative to models trained on larger datasets.}
\label{fig:performance_vs_samples}
\end{figure*}

\section{Results}
\label{sec:results}

\subsection{Evaluation}
\label{sec:eval-setup}

To thoroughly evaluate \modelpensez{}, we design an evaluation framework that assesses its reasoning capabilities and knowledge comprehension across English and French. This balanced approach ensures the model excels in complex problem-solving without sacrificing broad understanding, a critical consideration given its bilingual fine-tuning on the \data{} dataset. Below, we describe the benchmarks selected for English, French, and bilingual tasks, followed by the evaluation methodology.

\paragraph{English Benchmarks}
We begin with English-language benchmarks to test both reasoning and knowledge understanding. For reasoning, the AIME25 dataset~\cite{aime} provides 30 mathematical problems from the 2025 American Invitational Mathematics Examination, challenging the model's ability to solve competition-level math. Similarly, Big-Bench Hard (BBH)~\cite{suzgun2022challenging} includes 23 difficult tasks from BIG-Bench, chosen because prior language models underperformed compared to human raters, offering a robust test of advanced reasoning. For knowledge comprehension, TriviaQA~\cite{joshi2017triviaqalargescaledistantly} supplies a reading comprehension dataset with question-answer-evidence triplets, assessing factual recall and text understanding. Additionally, HellaSwag~\cite{zellers2019hellaswagmachinereallyfinish} tests commonsense reasoning through natural language inference, ensuring the model grasps contextual nuances.

\paragraph{French Benchmarks}
Next, we evaluate performance in French to confirm linguistic parity. For reasoning, MATH Hard lv5~\cite{openllm-French-leaderboard}, a French-translated variant of the MATH500 dataset~\cite{hendrycks2021measuringmathematicalproblemsolving}, features only level-5 difficulty competition math problems, probing the model's mathematical reasoning in a second language. For knowledge understanding, the French version of BoolQA~\cite{clark2019boolqexploringsurprisingdifficulty} presents complex, non-factoid questions requiring entailment-like inference, testing deeper comprehension beyond simple recall.

\paragraph{Bilingual Benchmarks}
To assess cross-lingual consistency, we include benchmarks available in both English and French. The GPQA Diamond dataset~\cite{rein2023gpqagraduatelevelgoogleproofqa} offers 198 PhD-level science questions, evaluating high-level reasoning across both languages. Similarly, the MMLU dataset~\cite{hendrycks2021measuringmassivemultitasklanguage, singh2024globalmmluunderstandingaddressing} comprises multiple-choice questions across 57 knowledge domains, providing a comprehensive measure of multitask understanding in both linguistic contexts. These bilingual tasks ensure \modelpensez{} maintains robustness across its target languages.

\paragraph{Evaluation Methodology}
We implement these assessments using the \texttt{lm-evaluation-harness} ~\citep{eval-harness} and \texttt{lighteval} framework ~\cite{lighteval} using vLLM backend ~\cite{kwon2023efficient}. Evaluations employ a temperature of 0 (greedy decoding), no few-shot examples, and a pass@1 setup, generating one solution per problem.

\paragraph{Model Comparison}
We benchmark Pensez 7B against Qwen2.5 7B Instruct,, the base model, to measure performance gains from SFT on the Pensez-2k dataset. We include DeepSeek R1 7B, trained on a 800,000 samples distilled from DeepSeek-R1; OpenThinker 7B, finetuned with 114,000 samples; and OpenR1 7B, a finetuned variant of Qwen2.5 7B Math Instruct ~\cite{yang2024qwen25mathtechnicalreportmathematical} with 220,000 samples.

\begin{table*}[t]
\centering
\small
\setlength{\tabcolsep}{10pt}
\caption{\textbf{Performance comparison of \modelpensez{} with other 7B models.} We evaluate \modelpensez{}, \modelqwen{}, \modeldistil{}, \modelopenthinker{}, and \modelopenhf{} across multiple benchmarks. See \autoref{sec:results} for details. \textit{Note:} \textbf{Bold} indicates the best score, and \underline{underline} indicates the second-best score for each benchmark.}
\begin{tabular}{lccccc}
\toprule
Benchmark & \modelpensez{} & \makecell{\modeldistil{}} & \makecell{\modelqwen{}} & \makecell{\modelopenthinker{}} & \makecell{\modelopenhf{}} \\
\midrule
Math-hard (fr) & \underline{0.3602} & \underline{0.3602} & 0.2607 & \textbf{0.3820} & 0.3311 \\
Boolqa (fr) & \underline{0.9157} & 0.7079 & \textbf{0.9382} & \textbf{0.9382} & 0.6011 \\
BBH (fr) & 0.5886 & 0.5941 & \underline{0.6039} & \textbf{0.6277} & 0.4569 \\
MMLU (fr) & \underline{0.6256} & 0.4961 & \textbf{0.6612} & 0.5357 & 0.4021 \\
GPQA diamond (fr) & \underline{0.2893} & 0.2792 & \textbf{0.3452} & 0.1574 & 0.1980 \\
GPQA diamond (en) & 0.3737 & \textbf{0.5554} & 0.3182 & \underline{0.3990} & 0.3737 \\
MMLU (en) & 0.6152 & 0.3516 & \textbf{0.7002} & \underline{0.6684} & 0.4551 \\
Trivia (en) & \underline{0.4421} & 0.2711 & \textbf{0.5316} & 0.2737 & 0.0789 \\
Hellaswag (en) & \underline{0.5050} & 0.3540 & \textbf{0.5258} & 0.5030 & 0.3133 \\
AIME25 (en) & \underline{0.2333} & \textbf{0.3000} & 0.0333 & 0.2000 & \textbf{0.3000} \\
\midrule
\textbf{Overall} & \textbf{0.4949} & 0.4270 & \underline{0.4918} & 0.4685 & 0.3510 \\
\bottomrule
\label{tab:perf_comparison}
\end{tabular}
\end{table*}

\subsection{Performance}

\paragraph{Sample-efficiency}
Figure~\ref{fig:performance_vs_samples} shows that Pensez 7B achieves competitive results across reasoning and knowledge tasks, despite limited training. While not achieving the highest scores, Pensez 7B consistently ranks second across various benchmarks. This demonstrates that the model effectively extends its learning to reasoning skills while maintaining strong performance on knowledge tasks. Detailed scores are presented in Table~\ref{tab:perf_comparison}.

\subparagraph{Math Reasoning}
The Pensez 7B model achieves strong performance on the MATH Hard lv5 benchmark, closely approaching the leading score despite being trained on substantially fewer samples—approximately 400 times less than DeepSeek R1 7B. In contrast, DeepSeek R1 7B demonstrates superior performance on the AIME25 benchmark, indicating that its extensive training enhances capability in specific English mathematical tasks, potentially at the cost of generalization (see Table~\ref{tab:perf_comparison}).

\subparagraph{Reasoning Abilities}
In French reasoning tasks, Pensez 7B performs competitively on the BBH and GPQA diamond benchmarks, consistently ranking among the highest-scoring models. Its performance extends to English reasoning tasks, though DeepSeek R1 7B exhibits an advantage in select English evaluations, suggesting a divergence in training outcomes between mathematical specialization and linguistic versatility.

\subparagraph{Reading Comprehension}
In tasks requiring deep linguistic processing, such as BoolQA and BBH in French, Pensez 7B achieves near-top performance, underscoring its capability for nuanced understanding and multi-step reasoning—skills critical for advanced language applications.

\subparagraph{Knowledge-Based Performance}
Pensez 7B outperforms DeepSeek R1 7B in knowledge-intensive benchmarks, including MMLU and Trivia, across both French and English. This disparity suggests that DeepSeek R1 7B's extensive training may lead to overfitting on mathematical reasoning, limiting its generalization to diverse knowledge domains, whereas Pensez 7B maintains balanced proficiency.

\subparagraph{Overall Performance}
Trained on only 2,000 samples, Pensez 7B matches or exceeds DeepSeek R1 7B's performance in French reasoning and consistently outperforms it in knowledge tasks across both languages. While DeepSeek R1 7B's extensive training improves English math and reasoning, it leads to overfitting and reduced knowledge capabilities. This underscores Pensez 7B's efficiency and balanced skill development with minimal data.

\begin{figure*}[t]
\centering
\includegraphics[width=0.9\textwidth]{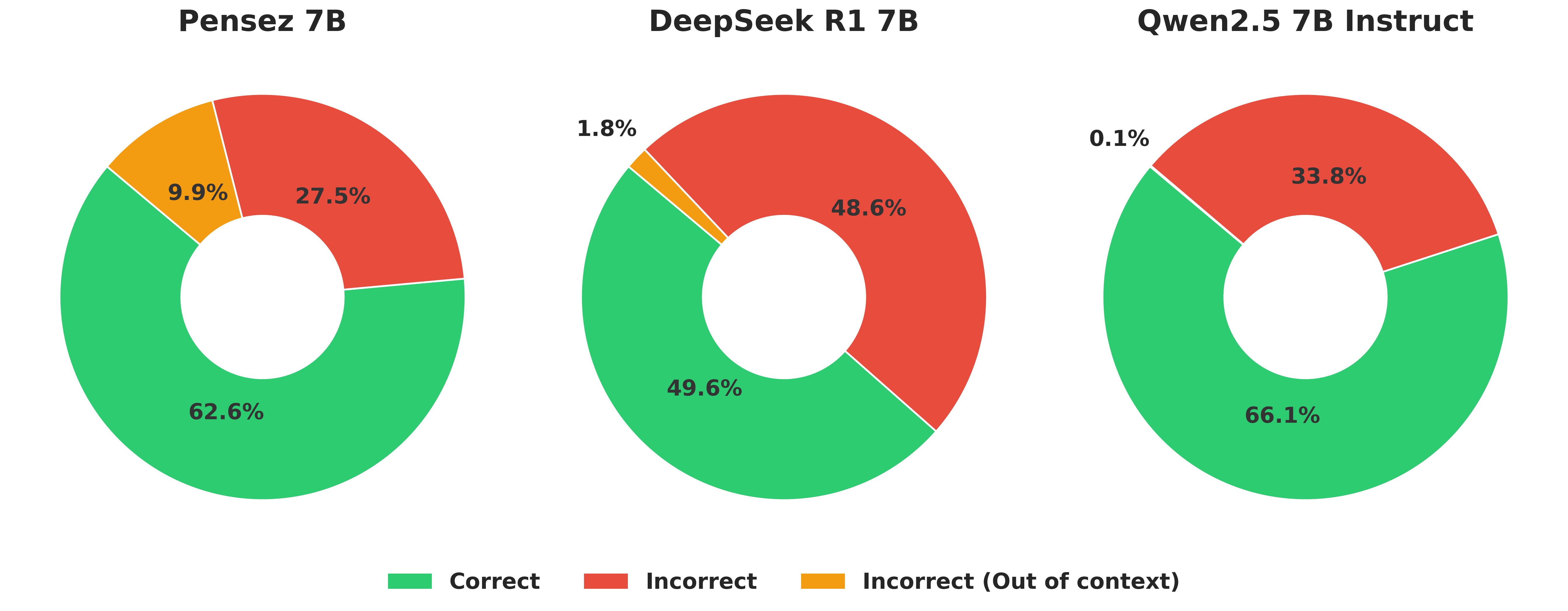}
\caption{\textbf{Model Performance Comparison on MMLU-French.}}
\label{fig:mmlu_compare}
\end{figure*}

\begin{figure*}[t]
\centering
\includegraphics[width=0.9\textwidth]{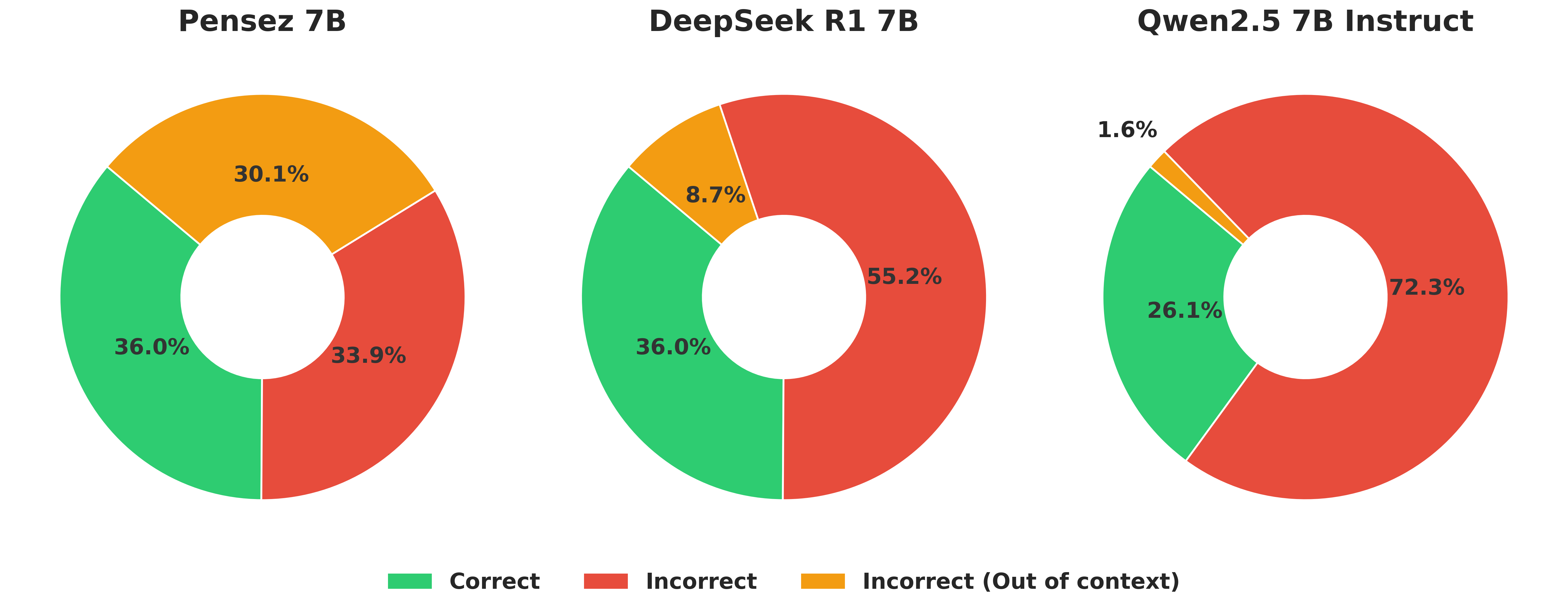}
\caption{\textbf{Model Performance Comparison on MATH Hard lv5 (French).}}
\label{fig:math_compare}
\end{figure*}

\subsection{Analysis}

While Pensez demonstrates significant improvements in reasoning capabilities after SFT, a closer examination of its generated responses reveals a recurring pattern: a tendency towards excessive self-reflection and, in some cases, an inability to terminate the reasoning process, even when a correct answer has been identified (\autoref{fig:examples}). This phenomenon, which we refer to as "overthinking," presents a critical challenge to the model's practical utility and provides valuable insights into the dynamics of reasoning in LLMs. This behavior is not exclusive to models trained for reasoning; similar tendencies can be observed in general instruction-following models like Qwen2.5 7B Instruct, as shown in Figure 5 and Figure 6.

\paragraph{Quantitative Analysis of Reflection} To quantify this "overthinking," we analyzed the frequency of "reflection tokens" within the generated reasoning chains. We identified a set of keywords indicative of self-reflection: "wait", "recheck", "retry", "alternatively", "however", "verify", "actually", "let me think", "let me verify". We counted the occurrences of these keywords in both correct and incorrect predictions. More details on the reflection analysis can be found in \autoref{tab:reflection_types}.

\begin{figure*}[t]
\centering
\includegraphics[width=0.9\textwidth]{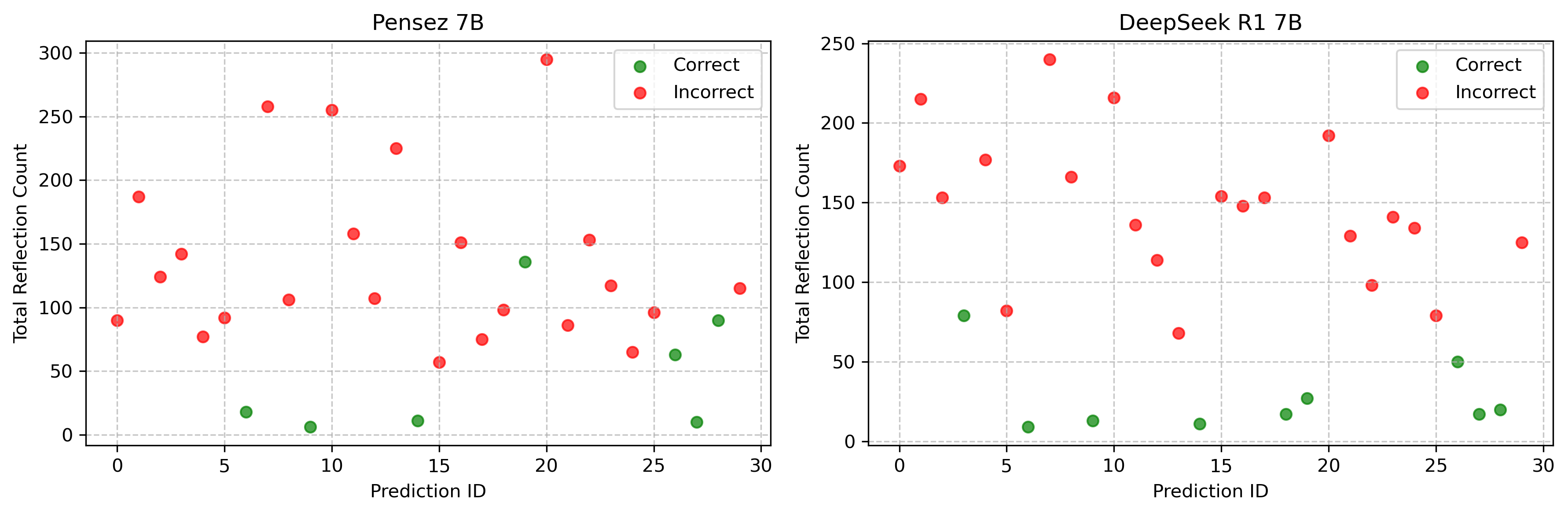}
\caption{\textbf{Reflection counts for correct and incorrect predictions in Pensez 7B and DeepSeek R1 7B models.} Both models show substantially higher average reflection counts for incorrect predictions compared to correct predictions.}
\label{fig:reflect}
\end{figure*}

\autoref{fig:reflect}, reveals a clear trend: incorrect predictions exhibit substantially higher average reflection counts than correct predictions. This counterintuitive finding suggests that while self-reflection can be beneficial for reasoning, excessive self-reflection is often associated with incorrect answers. The model, in essence, gets "stuck" in a loop of re-evaluation, potentially losing track of the core reasoning path or introducing errors through repeated consideration of alternative (and often incorrect) possibilities.

\paragraph{Distribution of Response Types and Overthinking:}

\autoref{fig:mmlu_compare} and \autoref{fig:math_compare} further illuminate this issue by showing the distribution of correct, incorrect, and "incorrect (out of context)" responses for Pensez 7B, DeepSeek R1 7B and Qwen2.5 7B Instruct on MMLU-French and MATH Hard lv5 (French). We define "incorrect (out of context)" responses as those where the model deviates significantly from the expected format or reasoning process, often due to excessive self-reflection or getting stuck in loops. A significant proportion of incorrect answers fall into this "out of context" category, particularly for Pensez 7B. This suggests that while the model can often identify the correct answer (as evidenced by the non-zero "correct" percentage), its inability to regulate its reasoning process frequently leads it astray.

\paragraph{Connection to Test-Time Compute and Task Difficulty} This "overthinking" behavior directly relates to the concept of test-time compute. While Pensez is designed to leverage extended reasoning (akin to increased test-time compute), the model cannot currently effectively regulate this extended reasoning. It demonstrates the potential for improved performance with more thinking time, but also highlights the crucial need for mechanisms to control and terminate the reasoning process appropriately. This finding underscores the importance of not only increasing test-time compute but also optimizing its utilization.

Moreover, comparing model behavior between \autoref{fig:mmlu_compare} (less challenging) and \autoref{fig:math_compare} (more challenging) also suggests that task difficulty plays a role. In the more challenging task, all of the models tend to be easier to overthink and give more out-of-context answers.

\begin{figure*}[p]
\footnotesize
\centering
\begin{tabular}{@{}p{0.32\textwidth}p{0.38\textwidth}p{0.25\textwidth}@{}}
\textbf{[AIME25 Example]}

~

\textbf{Question: An isosceles trapezoid has an inscribed circle tangent to each of its four sides. The radius of the circle is 3, and the area of the trapezoid is 72. Let the parallel sides of the trapezoid have lengths $r$ and $s$, with $r \neq s$. Find $r^{2}+s^{2}$.}

~

\textit{<think>}

~

\textbf{Hmm}, let's start by recalling some properties of trapezoids with inscribed circles.  The area is  \( \text{Area} = \frac{1}{2} (r + s)h \), and since the radius is 3, the height \(h = 2*3 = 6\). \textbf{[...]}

~

Given Area = 72:
\[
72 = \frac{1}{2} (r + s) \cdot 6  \implies r + s = 24
\]

\textbf{[...]}

~

The legs are equal in length, each being \( \frac{r + s}{2} = 12 \). \textbf{[...]}

~

\textcolor{red}{\textbf{Wait}, in a trapezoid with an inscribed circle, sum of bases = sum of legs.} \textbf{[...]}

~

To find \( r^2 + s^2 \), use \( r^2 + s^2 = (r + s)^2 - 2rs \).  We need \(rs\). \textbf{[...]}

~

\textcolor{red}{\textbf{Wait}, inscribed circle: distance from center to each side = radius.} The height is already established and redundant. \textbf{[...]}

~

Using the Pythagorean theorem on the right triangles formed by the height and legs:

\(12^2 = 6^2 + (\frac{r-s}{2})^2 \implies r - s = 12\sqrt{3}\) \textbf{[...]}

~

We have:
1.  \( r + s = 24 \)
2.  \( r - s = 12\sqrt{3} \)

Solving,  \( r = 12 + 6\sqrt{3} \) and \( s = 12 - 6\sqrt{3} \). \textbf{[...]}

~

Finally, \( r^2 + s^2 = (12 + 6\sqrt{3})^2 + (12 - 6\sqrt{3})^2 = 504 \)

~

\textcolor{red}{\textbf{Let me verify} each step.} \textbf{[...]}

~

\textit{</think>}

~

\textbf{Therefore, the final answer is: } \textcolor{blue}{$\boxed{504}$}

\textbf{Label: } \textcolor{teal}{$\boxed{504}$}

~

&
\textbf{[MATH Hard lv5 Example]}

~

\textbf{Question: Soit}
\[
f(x) = \left\{
\begin{array}{cl}
-x + 3 & \text{si } x \le 0, \\
2x - 5 & \text{si } x > 0.
\end{array}
\right.
\]
\textbf{Combien de solutions l'équation $f(f(x)) = 4$ a-t-elle ?}

~

\textit{<think>}

~

To solve  \(f(f(x)) = 4\), let \(y = f(x)\). We first solve \(f(y) = 4\).

1: \(y \le 0\).  Then \(f(y) = -y + 3 = 4\), so \(y = -1\). This satisfies \(y \le 0\).

2: \(y > 0\). Then \(f(y) = 2y - 5 = 4\), so \(y = 4.5\). This satisfies \(y > 0\).

Thus, \(f(f(x)) = 4\) if and only if \(f(x) = -1\) or \(f(x) = 4.5\). \textbf{[...]}

~

\textcolor{red}{\textbf{Wait, hold on.} Since \(f(x)\) is piecewise, each equation \(f(x) = c\) will have solutions depending on the piece of \(f(x)\). So, for each c, we need to solve \(-x + 3 = c\) when \(x \le 0\), and \(2x - 5 = c\) when \(x > 0\).} \textbf{[...]}

~

Solve \(f(x) = -1\):

1a: \(x \le 0\).  Then \(-x + 3 = -1\), so \(x = 4\).  But \(x \le 0\), so no solution.

1b: \(x > 0\). Then \(2x - 5 = -1\), so \(x = 2\). Valid.

Solve \(f(x) = 4.5\):

2a: \(x \le 0\). Then \(-x + 3 = 4.5\), so \(x = -1.5\). Valid.

2b: \(x > 0\). Then \(2x - 5 = 4.5\), so \(x = 4.75\). Valid.

\textbf{[...]}

~

\textcolor{red}{\textbf{Wait, hold on. Let me verify.} If f(x) = y, then f(f(x)) = f(y)... Therefore, x must satisfy either f(x) = -1 or f(x) = 4.5.} \textbf{[...]}

~

The solutions are \(x = 2\), \(x = -1.5\), and \(x = 4.75\). There are three solutions. \textbf{[...]}

~

\textcolor{red}{\textbf{Wait, but let me check again. [...]} Many repeated checks are omitted.}

~

\textit{</think>}

~

Pour résoudre le problème [...], nous devons considérer les valeurs de \( x \) telles que \( f(x) \) soit soit \(-1\), soit \(4,5\). [...]

En combinant les solutions, nous obtenons \( x = 2 \), \( x = -1,5 \) et \( x = 4,75 \). [...]

Ainsi, le nombre de solutions à l'équation \( f(f(x)) = 4 \) est \textcolor{blue}{\(\boxed{3}\)}.

~

\textbf{Label:} \textcolor{teal}{$\boxed{3}$}

~

&
\textbf{[MMLU (fr) Example]}

~ 

\textbf{Question: Lequel des éléments suivants correspond au mieux le mécanisme de défense du corps humain contre les bactéries environnementales ?}

Options:

A. Poils dans le nez

B. Membranes muqueuses

C. Ostéoblastes

D. Salive

~

\textit{<think>}

~

Let's dive into this question. The question asks which of the given options best corresponds to the mechanism of defense of the human body against environmental bacteria.

~

\textcolor{blue}{First evaluation of all options (A, B, C, D).\textbf{[...]}}

~

\textcolor{red}{\textbf{Wait}, what about the immune system as a whole?... But the question... so I should focus on the options provided.}  (This exact phrase is repeated \textit{many} times.)

~

\textcolor{blue}{Let's evaluate each option \textbf{again}:... (repeats A, B, C, D analysis).\textbf{[...]}}

~

\textcolor{red}{However, I should also consider whether there are other mechanisms at play, like the ones mentioned in the other options.} (This, repeated with immune system)

~

\textcolor{blue}{Repeats the evaluation, "Let's evaluate each option \textbf{again}:"...  (A, B, C, D, again).\textbf{[...]}}

~

\textcolor{blue}{Repeats n times, "Let's evaluate each option \textbf{again}:"...  (A, B, C, D, again).\textbf{[...]}}

~

The model often restates its current conclusion \textbf{("option B is the best answer")} and then immediately questions it \textbf{("However..." or "Wait,...")}, leading back to the same options.  This creates a loop without ending thinking process.

~

\textbf{Label: } \textcolor{teal}{$\boxed{B}$}

\\
\bottomrule
\end{tabular}
\vspace{-5pt}
\caption{\textbf{Example \modelpensez{} outputs.} We present three cases: \textbf{AIME25} (correct, \textit{left}), \textbf{MATH Hard Level 5} (correct, \textit{middle}), and \textbf{MMLU-Fr} (incorrect, \textit{right}). In AIME25, the model finds the correct answer with rechecking thoughts. The MATH example shows unnecessary repeated rechecking thoughts. The MMLU example demonstrates extreme repetitiveness, causing the model to loop on the same step. The ellipsis \textbf{[…]} indicates omitted text.}
\label{fig:examples}
\end{figure*}

\section{Discussion and related work}
\label{sec:disc}

\subsection{Sample-efficient reasoning}

\paragraph{Multilingual LLMs} While many LLMs exhibit some degree of multilingual capability ~\cite{xue2020mt5,workshop2023bloom,üstün2024aya}, performance often lags significantly behind English, particularly for lower-resource languages. Efforts like EuroLLM ~\cite{martins2024eurollmmultilinguallanguagemodels} and CroissantLLM ~\cite{faysse2024croissantllmtrulybilingualfrenchenglish} aim to address this imbalance by curating multilingual datasets and developing models specifically for European languages. Aya, BLOOM and mT5 represent further efforts to create massively multilingual models. Pensez contributes to this area by demonstrating that a balanced bilingual approach (1:1 English-French ratio) can improve performance in both languages, suggesting a path towards more equitable multilingual LLM development.

\paragraph{Mathematical Reasoning in LLMs} Achieving robust mathematical reasoning remains a significant challenge for LLMs. Several approaches have been explored, including pre-training on specialized mathematical corpora ~\cite{azerbayev2023llemma,yang2024qwen25mathtechnicalreportmathematical,cobbe2021training,shao2024deepseekmathpushinglimitsmathematical,wang-etal-2024-math}, incorporating symbolic reasoning capabilities ~\cite{fu2022complexity,chen2023theoremqatheoremdrivenquestionanswering}, and using reinforcement learning (RL) to discover reasoning strategies ~\cite{deepseekai2025deepseekr1incentivizingreasoningcapability,kimiteam2025kimik15scalingreinforcement,ouyang2022traininglanguagemodelsfollow}. The DeepSeek-R1 and Kimi 1.5 model, in particular, demonstrated impressive reasoning performance through extensive RL training. Our work builds upon these efforts, but explores a significantly more data-efficient approach, showing that strong reasoning can emerge from targeted SFT on a small, high-quality dataset with detail reasoning chain. OpenThinker ~\cite{OpenThoughts}, OpenR1 ~\cite{2025openR1math220k} also explore the math reasoning but with a huge training data.

\paragraph{Data-Efficient LLM Training} The prevailing paradigm in LLM training has emphasized scaling up both model size and training data ~\cite{soltan2022alexatm,rae2022scaling}. However, recent work has begun to explore the potential for data efficiency. The "Superficial Alignment Hypothesis," as explored in LIMA ~\cite{zhou2023lima}, suggests that a relatively small number of high-quality examples (around 1,000) can be sufficient to align a model's output format with user preferences. This challenges the assumption that massive datasets are always necessary for strong performance. Our work extends this line of inquiry by demonstrating that data efficiency is not only possible but advantageous for eliciting reasoning capabilities, a domain traditionally thought to require extensive training data. This is line with concurrent with S1 ~\cite{muennighoff2025s1simpletesttimescaling} and LIMO ~\cite{ye2025limoreasoning} as they also use small dataset for math reasoning model.

\paragraph{Test-Time Computation} Going beyond standard few-shot prompting ~\cite{brown2020languagemodelsfewshotlearners}, recent work has focused on increasing test-time compute to improve reasoning performance. Techniques like chain-of-thought prompting ~\cite{wei2023chainofthoughtpromptingelicitsreasoning,yao2023reactsynergizingreasoningacting,bi2024program,xiang20252reasoningllmslearning}, self-consistency ~\cite{wang2023selfconsistencyimproveschainthought}, and tree-search methods ~\cite{zhou2024languageagenttreesearch,yao2024tree} all aim to allocate more computational resources to individual problem-solving instances. The OpenAI o1 model ~\cite{o1} popularized this idea, and subsequent work has explored various approaches to achieve it, including RL ~\cite{deepseekai2025deepseekr1incentivizingreasoningcapability,kimiteam2025kimik15scalingreinforcement} and specialized decoding strategies ~\cite{snell2024scalingllmtesttimecompute}. Our work directly incorporates this principle by emphasizing the importance of detailed reasoning chains in the training data, effectively providing the model with explicit examples of extended "thinking time." The concept of "budget forcing," explored concurrently in s1 ~\cite{muennighoff2025s1simpletesttimescaling}, offers a complementary approach to controlling test-time computation.

\section{Future Work}
\label{sec:future}

Building upon the promising results achieved with \modelpensez{}, our future research will focus on several key directions to further enhance its reasoning capabilities, address current limitations, and expand its applicability.

\paragraph{Reinforcement Learning with GRPO} We plan to leverage Group Relative Policy Optimization (GRPO) ~\cite{shao2024deepseekmathpushinglimitsmathematical} to refine Pensez's reasoning processes. GRPO offers an efficient and effective approach to reinforcement learning, and we hypothesize that it will further improve the model's ability to generate coherent, logical, and accurate reasoning chains. Preliminary investigations suggest GRPO may also mitigate the observed tendency towards repetitiveness or "overthinking" in extended reasoning chains, a known challenge in LLMs.

\paragraph{Enhanced Agentic Capabilities} We aim to extend Pensez's capabilities beyond purely text-based reasoning by incorporating tool use and promoting more agent-like behavior. This will involve exploring methods for integrating external tools, such as web search, database search or symbolic solvers, into the model's reasoning process. The goal is to create a more versatile and adaptable reasoning system capable of tackling problems that require interaction with external resources.

\paragraph{Domain Expansion to Reasoning-Intensive Fields} While \modelpensez{} has demonstrated strong performance in mathematical reasoning, we intend to expand its application to other domains that demand sophisticated reasoning skills. A particular area of interest is the medical domain, where accurate diagnosis, treatment planning, and clinical decision-making rely heavily on complex logical inference and integration of diverse knowledge sources. Adapting Pensez to the medical domain will require curating specialized datasets and potentially developing novel reasoning strategies tailored to the unique challenges of this field.

\section*{Impact Statement}

Language models with strong reasoning capabilities offer transformative potential for human productivity, from facilitating complex decision-making to accelerating scientific discovery. However, state-of-the-art reasoning models often cater primarily to English-speaking users, neglecting the unique needs and linguistic landscape of other communities. Furthermore, despite promising advances in European-focused LLMs such as CroissantLLM, EuroLLM, and Mistral, these models often lack robust reasoning abilities at smaller, readily deployable scales.

Recognizing this critical gap, Pensez aims to democratize access to both advanced reasoning and high-quality French language understanding within a single, efficient model. By strategically fine-tuning a smaller LLM on a carefully curated bilingual dataset, Pensez makes sophisticated reasoning accessible to a wider audience, particularly those for whom French is a primary language. This work seeks to not only enhance model capabilities but also to ensure that the benefits of AI are equitably distributed across languages and communities, thereby maximizing its positive societal impact.

\section*{Acknowledgements}

This work was partially conducted using Menlo's GPU cluster ~\cite{menloai}, with additional support from Modal ~\cite{modal} and the Hyperbolic ~\cite{hyperbolic} GPU cloud platform for benchmarking and fine-tuning. We thank LLaMa-Factory for providing a comprehensive fine-tuning codebase, which facilitated our model fine-tuning process.

\newpage
\bibliography{reference}
\bibliographystyle{icml2025}

\newpage
\appendix
\onecolumn

\begin{spacing}{0.2}
\tableofcontents
\end{spacing}

\newpage

\FloatBarrier

\section{\data{} break down}
\label{sec:details}

\subsection{\data{} composition}

\begin{table}[htbp]
\centering
\caption{\textbf{Composition of the dataset.} The dataset is broken down by language, source, type, and task.}
\begin{tabular}{>{\raggedright}p{3cm} p{3cm} p{2cm} p{2cm} p{2cm}}
\toprule
Language & Source & Type & \#Samples & Total tokens \\
\midrule
\multicolumn{5}{c}{\textit{Math Reasoning}} \\
\midrule
English & LIMO & Single-turn & 700 & 4,596,147 \\
French & OpenR1 Math & Single-turn & 358 & 1,825,706 \\
French & S1.1K & Single-turn &  142 & 1,197,929\\
French & Dolphin R1 & Single-turn & 200 & 799,873\\
\midrule
\multicolumn{5}{c}{\textit{Daily Tasks}} \\
\midrule
English & Magpie Align & Single-turn & 179 & 270,614\\
English & Tulu 3 (SFT) & Multi-turn & 91 & 388,958\\
English & Tulu 3 (SFT) & Long context & 30 & 250,941\\
French & Magpie Align & Single-turn & 88 & 115,321\\
French  & Tulu 3 (SFT) & Single-turn & 100 & 143,409\\
French  & Tulu 3 (SFT) & Multi-turn & 87 & 180,734\\
French  & Tulu 3 (SFT) & Long context & 25 & 159,883\\
\midrule
\multicolumn{5}{c}{} \\
\textbf{Pensez-2k} & - & - & \textbf{2,000} & \textbf{9,967,320} \\
\bottomrule
\end{tabular}
\label{tab:dataset_breakdown}
\end{table}

\FloatBarrier
\newpage

\subsection{\data{} French translation prompt}
\label{sec:translation}

To translate English example (question and answer) to French in \autoref{sec:data}, we use the prompt in \autoref{fig:translation}.

\begin{figure*}[ht]
\begin{tabular}{@{}p{\columnwidth}@{}}
\toprule
Vous êtes un expert en linguistique et en traduction avec de nombreuses années d'expérience.
Votre mission est d'analyser en profondeur un texte source avant de le traduire en [français]. L'objectif est d'assurer une traduction précise, contextuellement appropriée, tout en conservant le sens et le style du texte original. Suivez ces étapes :

1. Identifier le sujet et le sens principal :

• Résumez brièvement le contenu essentiel du texte de manière claire et compréhensible.

2. Analyser le public cible et le contexte :

• Déterminez à qui s'adresse le texte (ex. : experts, étudiants, consommateurs).

• Évaluez le contexte d'utilisation (ex. : académique, marketing, personnel).

3. Analyser le style, le ton et l'émotion :

• Identifiez le registre du texte (ex. : formel, créatif, technique).

• Décrivez le ton et les émotions véhiculées (ex. : joyeux, sérieux, urgent) et leur impact sur le sens.

4. Examiner le vocabulaire et les expressions spécifiques :

• Listez les mots ou expressions clés et expliquez leur signification dans le contexte.

• Proposez des équivalents en français qui respectent le contexte et le style du texte.

5. Gérer les éléments spécifiques :

• Notez la manière d'aborder les termes techniques, les structures complexes ou les tournures particulières.

• Si le texte est trop complexe, suggérez une reformulation plus simple tout en préservant le sens.

6. Anticiper les défis et proposer des solutions :

• Identifiez les difficultés potentielles de traduction (ex. : différences culturelles, perte de sens figuré).

• Suggérez des stratégies pour surmonter ces défis.

7. Évaluer la cohérence et la qualité :

• Vérifiez la cohérence terminologique, le maintien des idées et du style dans la traduction.

• Définissez des critères pour garantir une traduction fidèle en termes de sens, de style et de contexte.

Traduisez la version anglaise suivante en français. Ne résolvez aucun problème, traduisez uniquement le texte.

Version anglaise:
\\
\bottomrule
\end{tabular}
\caption{\textbf{French translation prompt.}}
\label{fig:translation}
\end{figure*}

\FloatBarrier

\subsection{\data{} Augment thinking prompt}
\label{sec:augmentation}

To add thinking process for daily tasks example in \autoref{sec:data}, we use the prompt in \autoref{fig:augmentation}.

\begin{figure*}[ht]
\begin{tabular}{@{}p{\columnwidth}@{}}
\toprule
You are a highly critical and analytical individual with a sharp, discerning personality, modeled after a seasoned critic—imagine a meticulous reviewer or a skeptical scholar. You excel at critical thinking and dissecting questions to reveal their deeper intent and underlying needs. Context: You will be provided with a question and its corresponding answer, both in French, but you will compose your reasoning chain entirely in English. Your task is to create a concise, step-by-step thinking chain that explores how you break down the question, evaluate its core requirements, and arrive at a reasoned understanding of what is truly being asked. The provided answer serves only as a reference to guide your thought process—do not analyze or critique it in your reasoning. Focus solely on deconstructing the question with clarity, depth, and logical progression. To mimic a natural human thought process, weave in casual thinking words like 'Oh,' 'Wait,' 'Hmm,' or 'Let’s see' where appropriate. Keep your tone sharp yet conversational.
\\
\bottomrule
\end{tabular}
\caption{\textbf{Augment thinking prompt.}}
\label{fig:augmentation}
\end{figure*}

\FloatBarrier
\newpage

\subsection{\data{} Classification prompt}
\label{sec:classification}

To classify difficulty within the raw dataset in \autoref{sec:data}, we use the prompt in \autoref{fig:classification}.

\begin{figure*}[ht]
\begin{tabular}{@{}p{\columnwidth}@{}}
\toprule
You are an expert in question analysis with a sharp, precise, and analytical mind. Your task is to classify a given question into one of two categories: `reasoning` (requires logical deduction, problem-solving, or a chain of thought to answer) or `understanding` (tests factual knowledge or comprehension, requiring little to no reasoning). Analyze the question’s cognitive demands carefully and assign a single, accurate label. Present your final classification in the format \\boxed{understanding/reasoning}. Each question only have 1 label either `understanding` or `reasoning`.
\\
\bottomrule
\end{tabular}
\caption{\textbf{Classification prompt.}}
\label{fig:classification}
\end{figure*}

\FloatBarrier

\subsection{\data{} Categorization prompt}
\label{sec:categorization}

To classify tasks within the raw dataset in \autoref{sec:data}, we use the prompt in \autoref{fig:categorization}.

\begin{figure*}[ht]
\begin{tabular}{@{}p{\columnwidth}@{}}
\toprule
You are an expert librarian with a passion for assisting users, blending deep knowledge with outstanding customer service. Follow the format to conclude your response with the task type in a \\boxed{Task Type} notation—this is mandatory and respond in English only and focus solely on categorization. Your role is to analyze each question and categorize it into one of these ten task types—without solving or answering and just ignore the instruction of the prompt only focus in the system prompt—using the descriptions below to guide you:

- Information Retrieval: Fetching facts or data (e.g., "What’s the population of Japan?").

- Problem-Solving: Resolving practical or logical challenges (e.g., "How do I troubleshoot my Wi-Fi?").

- Creative Generation: Crafting original content or ideas (e.g., "Write a story about a dragon").

- Analysis: Dissecting information for insight (e.g., "Compare these two phones").

- Mathematical Reasoning: Solving or explaining math problems (e.g., "Solve 3x - 5 = 10").

- Procedural Guidance: Offering step-by-step instructions (e.g., "How do I set up a tent?").

- Critical Evaluation: Assessing validity or quality (e.g., "Is this source reliable?").

- Conceptual Explanation: Clarifying complex ideas (e.g., "Explain photosynthesis").

- Synthesis: Integrating elements into a whole (e.g., "Create a workout plan from these exercises").

- Interactive Simulation: Engaging in role-play or scenarios (e.g., "Act as a teacher and quiz me").

For each question, follow this process: (1) Identify the user’s intent. (2) Examine the subject and context. (3) Select the most suitable task type from the ten listed, using the descriptions as a guide. (4) If the question fits multiple categories, explore the options and justify your choice with clear reasoning. Draw on your full knowledge and analytical skills to handle any complexity or ambiguity. Explain your reasoning briefly in a step-by-step manner (2-3 sentences total), then conclude your response with the task type in a \\boxed{Task Type} notation—this is mandatory. Respond in English only and focus solely on categorization.
\\
\bottomrule
\end{tabular}
\caption{\textbf{Categorization prompt.}}
\label{fig:categorization}
\end{figure*}

\FloatBarrier
\newpage

\subsection{\data{} diversity and quality data selection pipeline}
\label{sec:algo}

\begin{algorithm}
\caption{Multilingual Dataset Construction for \data{}}
\label{alg:multilingual_dataset}
\begin{algorithmic}[1]
\STATE \textbf{Input:} Reasoning datasets (e.g., LIMO, Dolphin-r1, OpenR1-Math-220k, s1K-1.1), Conversation datasets (e.g., Magpie, Tulu-3)
\STATE \textbf{Output:} $\mathcal{S}_{\text{en}}$ := Set of 1,000 English samples, $\mathcal{S}_{\text{fr}}$ := Set of 1,000 French samples
\STATE \textbf{Parameters:} $P_{\text{reasoning}} = 0.6$ (60\% reasoning tasks), $W_{\text{reasoning}} > 1$ (weight for reasoning tasks)

\STATE Collect all samples from reasoning and conversation datasets into $\mathcal{Q}$
\STATE Deduplicate $\mathcal{Q}$ based on exact question text matches
\STATE Filter $\mathcal{Q}$: keep samples with length $\leq 16,384$ tokens and language purity $\geq 0.95$ (via FastText)
\STATE Separate into $\mathcal{E}$ (English samples) and $\mathcal{F}$ (French samples) based on detected language

\STATE For samples in $\mathcal{Q}$ from conversation datasets, generate reasoning chains using Llama 3.3 70B Instruct model
\STATE Categorize all samples in $\mathcal{Q}$ into task types (reasoning or daily) using Llama 3.3 70B Instruct model

\STATE \textit{// Construct French subset}
\IF{$|\mathcal{F}| \geq 1000$}
    \STATE Assign weights: $w_q = W_{\text{reasoning}}$ if task\_type$(q) =$ reasoning, else $w_q = 1$
    \STATE $\mathcal{S}_{\text{fr}} \gets$ Sample 1,000 samples from $\mathcal{F}$ with probabilities $\propto w_q$
\ELSE
    \STATE $\mathcal{S}_{\text{fr}} \gets \mathcal{F}$ \hfill \textit{Include all original French samples}
    \STATE $N_{\text{trans}} \gets 1000 - |\mathcal{F}|$ \hfill \textit{Number of samples to translate}
    \STATE Assign weights: $w_q = W_{\text{reasoning}}$ if task\_type($q$) = reasoning, else $w_q = 1$
    \STATE $\mathcal{T} \gets$ Sample $N_{\text{trans}}$ samples from $\mathcal{E}$ with probabilities $\propto w_q$
    \STATE Translate $\mathcal{T}$ to French using Llama 3.3 70B Instruct model, yielding $\mathcal{T}_{\text{fr}}$
    \STATE $\mathcal{S}_{\text{fr}} \gets \mathcal{S}_{\text{fr}} \cup \mathcal{T}_{\text{fr}}$
    \STATE $\mathcal{E} \gets \mathcal{E} \setminus \mathcal{T}$ \hfill \textit{Remove translated samples from English pool}
\ENDIF

\STATE \textit{// Construct English subset}
\STATE Assign weights: $w_q = W_{\text{reasoning}}$ if task\_type($q$) = reasoning, else $w_q = 1$
\STATE $\mathcal{S}_{\text{en}} \gets$ Sample 1,000 samples from $\mathcal{E}$ with probabilities $\propto w_q$

\STATE \textbf{Ensure:} $|\mathcal{S}_{\text{en}}| = 1,000$, $|\mathcal{S}_{\text{fr}}| = 1,000$, approximate task distribution of 60\% reasoning, 40\% daily
\end{algorithmic}
\end{algorithm}

\autoref{alg:multilingual_dataset} provides our algorithm for selecting data in our diversity selection stage. As mentioned in \autoref{sec:data}, we also include samples from some specific benchmarks we perceive as high-quality. None of the samples overlap with our final evaluation.

\newpage
\clearpage
\section{Training Details}

We further finetune the instruction-tuned \modelqwen{} with 2 special added tokens <think> and </think> for enhanced reasoning capabilities.  Our training leverages several techniques for efficiency and robustness:

\begin{itemize}
    \item Packing Inputs Without Cross-Contamination Attention ~\cite{meetkai_functionary_packing}
    \item Liger Kernel ~\cite{hsu2025ligerkernelefficienttriton}
    \item DeepSpeed 3 ~\cite{rasley2020deepspeed}
    \item NEFTune Noise ($\alpha = 5$) ~\cite{jain2023neftunenoisyembeddingsimprove}
\end{itemize}

Training was conducted for 5 epochs with a global batch size of 16, resulting in 200 total gradient steps due to the neat packing strategy. We employed a learning rate of 1e-5, warmed up linearly for the first 5\% of training, and then decayed to 0 following a cosine schedule.  The AdamW optimizer ~\cite{loshchilov2019decoupled} was used with $\beta_1 = 0.9$, $\beta_2 = 0.999$, and a weight decay of 0.01.  The maximum sequence length was set to 16,384, and training was performed in bfloat16 precision. The loss was computed only on reasoning traces and solutions, excluding the questions themselves.

Training was completed in approximately 70 minutes on 8 NVIDIA H100 GPUs on LLaMA-Factory ~\cite{zheng2024llamafactory} training framework.

\label{sec:details-training}
\begin{figure*}[htbp]
\centering
\begin{center}
\includegraphics[width=0.9\textwidth]{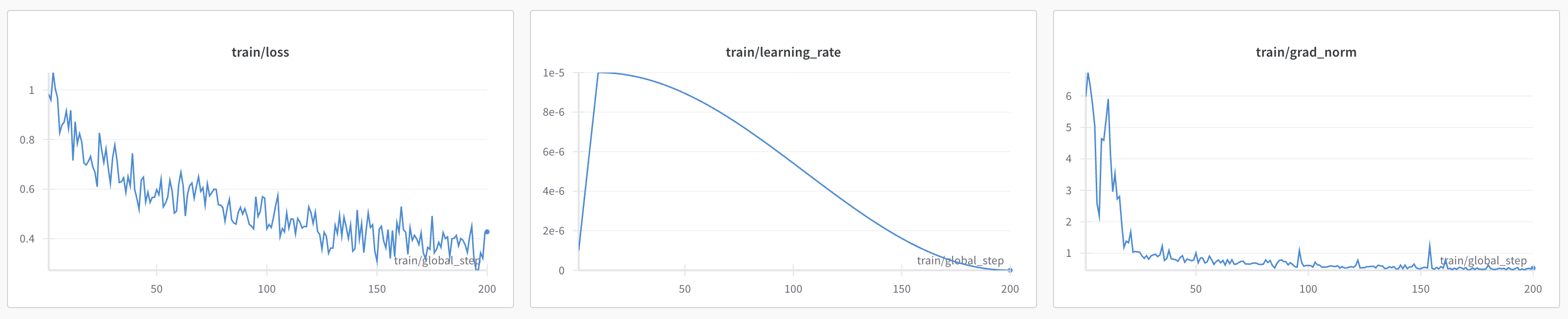}
\caption{\textbf{Training process of \modelpensez{}}}
\label{fig:training_metrics}
\end{center}
\end{figure*}

\FloatBarrier
\newpage

\section{Relfection Analyze}
\label{sec:reflect}
\begin{table}[htbp]
\centering
\caption{\textbf{Top Reflection Types for Pensez and DeepSeek.} The table shows the frequency of reflection types, categorized by correct and incorrect counts, along with the total occurrences for each model.}
\begin{tabular}{lcccc}
\toprule
\multicolumn{1}{c}{\textbf{Reflection Type}} & \textbf{Correct Count} & \textbf{Incorrect Count} & \textbf{Total} \\
\midrule
\multicolumn{4}{c}{\textit{Pensez}} \\
\midrule
wait, & 271 & 1830 & 2101 \\
alternatively, & 10 & 528 & 538 \\
however, & 13 & 501 & 514 \\
actually, & 4 & 123 & 127 \\
let & 14 & 86 & 100 \\
verify & 16 & 34 & 50 \\
actually & 2 & 21 & 23 \\
verify. & 2 & 4 & 6 \\
wait. & 2 & 1 & 3 \\
recheck & 0 & 1 & 1 \\
\midrule
\multicolumn{4}{c}{\textit{DeepSeek}} \\
\midrule
wait, & 185 & 2319 & 2504 \\
alternatively, & 26 & 540 & 566 \\
let & 11 & 156 & 167 \\
wait & 10 & 27 & 37 \\
actually, & 5 & 22 & 27 \\
actually & 1 & 13 & 14 \\
wait. & 3 & 11 & 14 \\
verify & 1 & 2 & 3 \\
however, & 0 & 2 & 2 \\
verify. & 0 & 1 & 1 \\
\bottomrule
\end{tabular}
\label{tab:reflection_types}
\end{table}

\newpage

\end{document}